\documentclass{article}
\usepackage{ijcai23}

\usepackage{times}
\usepackage{soul}
\usepackage{url}
\usepackage[hidelinks]{hyperref}
\usepackage[utf8]{inputenc}
\usepackage[small]{caption}
\usepackage{algorithm}
\usepackage{algorithmic}
\usepackage[switch]{lineno}
\usepackage{graphicx}
\usepackage{amsmath}
\usepackage{amsthm}
\usepackage{caption}
\usepackage{subcaption}
\usepackage{booktabs}
\usepackage{algorithm}
\usepackage{algorithmic}
\usepackage{amsfonts,amssymb}
\usepackage{color}
\usepackage{enumerate}
\usepackage{bm}
\usepackage{multirow}
\usepackage{arydshln}
\newtheorem{definition}{Definition}

\usepackage{pifont}

\title{FedCliP: Federated Learning with Client Pruning}

\author{
Beibei Li$^1$
\and
Zerui Shao$^1$\and
Ao Liu$^{1}$\And
Peiran Wang$^2$
\affiliations
$^1$School of Cyber Science and Engineering, Sichuan University, Chengdu, China\\
$^2$Institute for Network Sciences and Cyberspace, Tsinghua University, Beijing, China
\emails
libeibei@scu.edu.cn,
shaozerui@stu.scu.edu.cn,
liuao@stu.scu.edu.cn,
wangpr22@mails.tsinghua.edu.cn
}

\begin{document}
\maketitle

\begin{abstract}\label{section.abstract}
      The prevalent communication efficient federated learning (FL) frameworks usually take advantages of model gradient compression or model distillation. However, the unbalanced local data distributions (either in quantity or quality) of participating clients, contributing non-equivalently to the global model training, still pose a big challenge to these works. In this paper, we propose FedCliP, a novel communication efficient FL framework that allows faster model training, by adaptively learning which clients should remain active for further model training and pruning those who should be inactive with less potential contributions. We also introduce an alternative optimization method with a newly defined $contribution\ score$ measure to facilitate active and inactive client determination. We empirically evaluate the communication efficiency of FL frameworks with extensive experiments on three benchmark datasets under both IID and non-IID settings. Numerical results demonstrate the outperformance of the porposed FedCliP framework over state-of-the-art FL frameworks, i.e., FedCliP can save 70\% of communication overhead with only 0.2\% accuracy loss on MNIST datasets, and save 50\% and 15\% of communication overheads with less than 1\% accuracy loss on FMNIST and CIFAR-10 datasets, respectively.
\end{abstract}

\section{Introduction}\label{section.Introduction}
    Federated learning (FL) generates an effective global model by training local models independently on each client and aggregating local model parameters on a central server while preserving data privacy \cite{fedavg,fedprox,ditto,scaffold,fedcur}. Despite the FL schemes' success, the FL training methods still suffer from high communication costs because clients are usually equipped with limited hardware resources and network bandwidth, which suspend the deployment of large neural networks on a wide scale. Communication efficiency is a significant bottleneck restricting the development and deployment of FL.

    Various works have been proposed to solve the communication bottleneck, and we can divide these methods into four categories: (1) Accelerated model convergence techniques \cite{power,afl,fedreg,fedcor,progfed} reduce the overall communication cost by reducing the number of communication rounds required for global model convergence, (2) compression techniques \cite{quantization,sparsification,fedpara,quantize} that utilize fewer bits to represent gradients or parameters to reduce communication costs, (3) model pruning techniques \cite{pruningfed,sparsefed} that identify much smaller sub-networks to reduce the communication overhead. And (4) knowledge distillation \cite{fedds,ensembledistillation} techniques allow the server to extract client knowledge.

    The works mentioned above achieve communication efficiency from the point of view of compressing model parameters or searching for critical parameters or vital structures of the client model. However, the unbalanced local data distributions of participating clients, contributing non-equivalently to the global model convergence, still pose a significant challenge to these works. When the client model or data does not contribute to the global model training, even if the above communication efficient methods are used, the overhead of model communication will still be incurred because these methods will not discard all the client model parameters. Our key idea is mainly based on the following intuitions:

    Clients do not contribute equivalently. Training clients with a large and balanced dataset can be more beneficial to the convergence of the global model, while training clients with small and highly biased datasets may increase the losses of the global model. The local model convergence rate is inconsistent, caused by data imbalance. A client participates in global training many times, the local model tends to converge gradually, and its contribution to the global model will decrease progressively. It will occupy valuable model training resources and harm model convergence if this client still participates in the training.

    This means we should pay more attention to those have more effect on global model convergence to participate in training, and clients with less contribution even play a negative role will be pruned gradually with the FL training. In this paper, we propose a client-pruning strategy, FedCliP, that can quickly identify essential active clients through several communication rounds and progressively reduce the inactive clients participating in FL training. The pruned inactive clients will no longer participate in the following training process. Screening essential clients during the FL training process narrows the scope of participating in training clients continuously, and achieves communication efficiency from the macro perspective of the FL framework. Our contributions in this paper are summarized as follows:

    \begin{itemize}
      \item We propose the first communication efficient FL framework with client pruning based on our knowledge, dubbed FedCliP, which can identify active clients that contribute more to the global model and minimize communication overhead by excluding inactive clients that contribute less to global model training from the FL training process.

      \item We define a client $contribution\ score$ measure to promote active and inactive client decisions. The client $contribution\ score$ can adaptively evaluate the contribution value of the client in this round, measure each client's data quality and model difference, and fully utilizes each client's resources.

      \item We introduce an alternative optimization method to determine the next round of active client sets. We utilize Gaussian Scale Mixture (GSM) modeling to model the client $contribution\ score$, efficiently estimate the numerical components of active clients, and then realize the screening of inactive clients.

      \item We have conducted numerous experiments on three benchmark datasets under IID and non-IID settings. The numerical results show that the proposed FedCliP can save 70\% of the communication overhead, only 0.2\% of the accuracy is lost on the MNIST dataset, and 50\% and 15\% of the communication overhead are saved, with less than 1\% of the accuracy loss on the FMNIST and CIFAR-10 datasets, respectively.
    \end{itemize}

\section{Related Works}\label{section.RelatedWorks}
\subsection{Federated Learning}
  Federated Learning (FL) enables participating clients to collaboratively train a model without migrating the client’s data, which addresses privacy concerns in a distributed learning environment \cite{fedbn,fedu,fola,fedper,lgfedavg,pfedhn,apple}. FedAvg \cite{fedavg} is the the most widely used FL algorithm. In FedAvg framework, the server sends the initial parameters to the participating clients. The parameters are updated independently on each of these clients to minimize the local data loss, and the locally trained parameters are sent to the server. The server aggregates the local parameters by simply averaging them and obtains the next round of global parameters. Many studies focus on amending the local model updates or the central aggregation based on FedAvg. FedProx  \cite{fedprox} is another notable FL framework that stabilizes FedAvg by including a proximal term in the loss function to limit the distance between local and global models. FedReg \cite{fedreg} proposed to encode the knowledge of previous training data learned by the global model and accelerate FL with alleviated knowledge forgetting in the local training stage by regularizing locally trained parameters with the loss on generated pseudo data.

\subsection{Communication Efficient Federated Learning}
  Fruitful literature has studied model compression to reduce communication costs in distributed learning. The basic idea of these methods is that before uploading the trained local model to the server, first compress and encode the model parameters or gradients, and then the communication is efficient by uploading a small amount of data. Compression on server-to-client communication is non-trivial and attracted many recent focuses \cite{fedpara,quantization,sparsification}. Pruning and distillation methods have been combined with federated learning to achieve model compression in recent years \cite{bip,prospr,pruningfed,fedds,ensembledistillation}. Model pruning removes inactive weights to address the resource constraints, and model distillation requires a student model to mimic the behavior of a teacher model. These methods aim to find important model parameters or structures and achieve communication efficiency by transferring essential parts of the model. In addition, the way to achieve efficient federated learning communication is to speed up the convergence of the model \cite{fedreg}. Because the convergence of the global model is faster, the required communication rounds will be reduced accordingly, and the communication costs will be reduced accordingly. However, they did not consider the impact of the unbalanced distribution of client data on the global model training. When the client model has no contribution to the global model training, the above methods cannot achieve real communication efficiency because they cannot discard all the local model parameters in principle. The FedCliP method proposed in this paper can adaptively prune the client during the model training process, and the pruned client cannot continue to participate in the federated learning process, thus reducing the communication overhead to the greatest extent.

\section{Methodology}\label{section.Methodology}
\begin{figure}[t]
	\centering
	\includegraphics[width=1\linewidth]{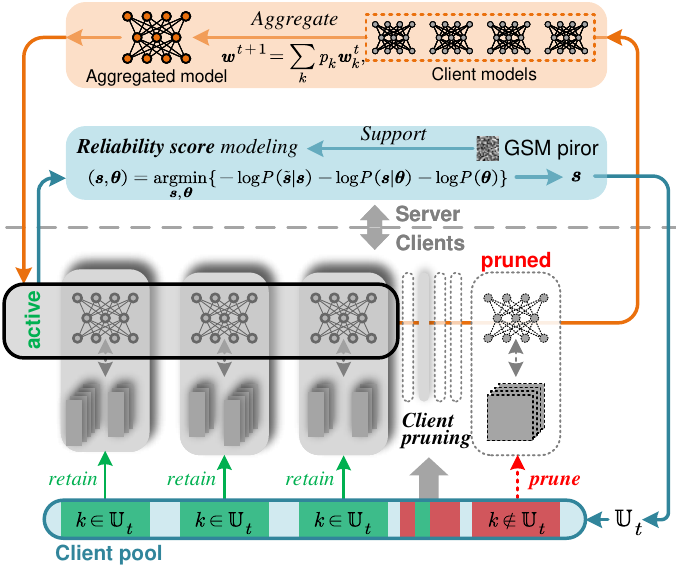}
	\caption{The framework of the proposed FedCliP. The server utilizes the GSM to model the $contribution\ score$ calculated by each client and updates the set of active clients participating in the next round of training.
	\label{figure.FedCliP}
	}
  \end{figure}

  In this section, we elaborate our proposed FL framework, i.e., FedCliP, that can achieve communication efficient. We first introduce the goal of our FedCliP framework in Section \ref{subsection.Formulation}. Then, the client $contribution\ score$ calculation method is given in Section \ref{subsection.Score}, based on the data quality and model difference. Then, we provide empirical evidence that the prior distribution of client $contribution\ score$ in each communication round can be modeled as Gaussian distribution. Based on this observation, we utilize GSM to model the client pruning problems in Section \ref{subsection.GSM} and obtain an effective client pruning optimization strategy for FL in Section \ref{subsection.Optimization}. We illustrate the framework of FedCliP in Figure \ref{figure.FedCliP}.

\subsection{Problem Formulation}\label{subsection.Formulation}
  In the FL, the client $k$ keeps its local data $\mathbb{D}_{k}$ on the client machine during the whole training process. FL aims to seeks for a global model $\boldsymbol{w}$ that achieves the best performance on all $N$ clients. We refer to $l_{k}(\boldsymbol{w})$ as the local loss of client $k$, which is evaluated with the local dataset $\mathbb{D}_{k}$, which size is $\left|\mathbb{D}_{k}\right|$, on client $k$. The weight $p_{k}=\left|\mathbb{D}_{k}\right| / \sum_{j}\left|\mathbb{D}_{j}\right|$ of the client $k$ is proportional to the size of its local dataset.

  In this paper, in communication round $t$, we select the overall clients in client set $\mathbb{U}_{t}$ to receive the global model $\boldsymbol{w}^{t}$ and conduct training with their local dataset for several iterations independently. After the local training, the server collects the trained models from these selected clients and aggregates them to produce a new global model $\boldsymbol{w}^{t+1}$. Our goal is to dynamically evaluate the $contribution\ score$ of each client during the FL process and determine whether the client will continue to participate in the next training round based on the $contribution\ score$. Clients who can continue participating in the following training round will constitute a new collection of clients $\mathbb{U}_{t}$ for $t+1$ communication rounds.

\subsection{Contribution Score}\label{subsection.Score}
  To effectively identify inactive clients, we measure each client’s model with the $contribution\ score$, which jointly considers the contribution of the local model and data quality to the global model. After the client receives the global model and completes local training, the difference between the current local model and the global model is an essential indicator for judging whether the client needs to continue to participate in the next round of training. A slight difference value means that the client data contributes less to the convergence of the global model, and the probability of the client continuing to participate in training is lower. Similarly, as the training proceeds, the local model of an active client and the updated global model has a relatively large distance. The model difference is an effective metric to quantify the model quality in FL. Based on this point of view, we define the model difference below.

  \begin{definition}
  We define the model difference $d$ for the $k$-th client as the $\ell_2$-norm distance between the current local model $\boldsymbol{w}_{k}^{t}$ and last-round global model $\boldsymbol{w}^{t-1}$, i.e.

  \begin{equation}\label{equation.modelrs}
  \small
      d_{k}^{t}=\left\|\boldsymbol{w}_{k}^{t}-\boldsymbol{w}^{t-1}\right\|_{2}^{2},
  \end{equation}

  \end{definition}

  As the most critical resource for the client, data quality is another important criterion for measuring the client's contribution. Clients with high-quality data often need to continue participating in the federated learning process because high-quality data will positively affect the convergence of the global model. However, data quality has always been challenging to evaluate and predict. To measure client data quality, we quantify data quality from two aspects: the convergence speed of the local model on the client data and the amount of data. We utilize the local model's training loss to measure the local model's convergence speed. The data size is often used to evaluate the client data quality, and we continue to use this point of view. Overall, we define the client’s data quality below.

  \begin{definition}
  We define the data quality $q$ for the $k$-th client using the client training loss of the local training data and the data size, i.e.

  \begin{equation}\label{equation.datars}
  \small
      q_{k}^{t}=\left|\mathbb{D}_{k}\right|\sqrt{\frac{1}{\mathbb{D}_{k}}\sum_{d\in\mathbb{D}_{k}}l_{k}(\boldsymbol{w}_{k}^{t},d)^{2}}.
  \end{equation}

  \end{definition}

  At last, we integrate the model difference of Eq. (\ref{equation.modelrs}) and local data quality of Eq. (\ref{equation.datars}) into our $contribution\ score$, and we define the $contribution\ score$ below.

  \begin{definition}
  We define the $contribution\ score$ $s$ for the $k$-th client using the model difference and local data quality, i.e.

  \begin{equation}\label{equation.clientrs}
  \small
      s_{k}^{t}=d_{k}^{t} * q_{k}^{t}.
  \end{equation}

  \end{definition}

  \noindent As an important indicator to judge whether the current client continues to participate in the federated learning process, the $contribution\ score$ makes full use of the information of the local model and data and plays a vital role in dynamically pruning the client.

\subsection{Gaussian Scale Mixture Modeling for Contribution Scores}\label{subsection.GSM}
  After getting the $contribution\ score$ of each client, we can pruning the inactive clients with $contribution\ scores$ $\boldsymbol{s}^{t}=\left\{s_{1}^{t}, \ldots s_{N}^{t}\right\}$ obtained in communication round $t$. We regard identifying inactive clients from the $contribution\ scores$ of all clients as a data-denoising task. By considering the $contribution\ scores$ of inactivate clients as Gaussian noise of clean $contribution\ scores$ of all clients, we utilize the GSM to model the $contribution\ scores$ and obtain the clean $contribution\ scores$ representing the active clients. For simplicity, we set $\Tilde{\boldsymbol{s}}\in\mathbb{R}^{N}$ and $\boldsymbol{s}\in\mathbb{R}^{N}$ denote the one-dimensional representations of all clients' $contribution\ scores$ and activate clients' $contribution\ scores$, respectively. Then $\boldsymbol{s}$ denotes the noiseless version of $\Tilde{\boldsymbol{s}}$, i.e., $\Tilde{\boldsymbol{s}}=\boldsymbol{s}+\boldsymbol{n}$, where $\boldsymbol{n}\in\mathbb{R}^{N}$ denotes additive Gaussian noise of inactive clients' $contribution\ scores$. In this subsection, we propose a Maximum A Posterior (MAP) method for estimating $\boldsymbol{s}$ from $\Tilde{\boldsymbol{s}}$, and the MAP estimation of $\boldsymbol{s}$ from $\Tilde{\boldsymbol{s}}$ can be formulated as

  \begin{equation}\label{equation.map}
  \small
      \boldsymbol{s}=\mathop{\arg\min}_{\boldsymbol{s}}\{-\log P(\Tilde{\boldsymbol{s}} | \boldsymbol{s})-\log P(\boldsymbol{s})\},
  \end{equation}

  \noindent where $\log P(\Tilde{\boldsymbol{s}} | \boldsymbol{s})$ is given by the Gaussian distribution of noise from inactive clients, i.e.,

  \begin{equation}
  \small
      P(\Tilde{\boldsymbol{s}} | \boldsymbol{s}) \propto \exp
      \left(-\frac{1}{2 \sigma_{w}^{2}}\|\Tilde{\boldsymbol{s}}-\boldsymbol{s}\|_{2}^{2}\right),
  \end{equation}

  \noindent where $\exp(\cdot)$ denotes the exponential operation on each element of the input parameters, and $\sigma_w^2$ is the variance of the Gaussian noise. A prior distribution of $\boldsymbol{s}$ is given by Laplacian, and it is easy to verify that the above MAP estimation leads to the following weighted $\ell_1$-norm minimization problem when $P(\boldsymbol{s})$ is chosen to be an IID Gaussian,

  \begin{equation}
  \small
      \boldsymbol{s}=\mathop{\arg\min}_{\boldsymbol{s}}
      \|\Tilde{\boldsymbol{s}}-\boldsymbol{s}\|_{2}^{2}+
      2 \sigma_{w}^{2} \sum_{j} \frac{1}{\theta_{j}}\left|s_{j}\right|,
  \end{equation}

  \noindent where $\theta_{j}$ denotes the standard derivation of $s_{j}$. In realistic scenarios, it is difficult to estimate the $\theta_{j}$ from $\Tilde{\boldsymbol{s}}$. The right side of Figure \ref{figure.ablation} shows the probability distribution of all client $contribution\ scores$ after completing the training. It can be seen that during the dynamic training process, the distribution of client $contribution\ scores$ roughly satisfies the Gaussian distribution. It can also be explained from the side of the rationality of our setting of $contribution\ score$ measure.

  In this paper, we propose a Gaussian Scale Mixture (GSM) prior for $\boldsymbol{s}$. GSM model has demonstrated its potential for signal modeling in computer vision works \cite{gsm02,gsm03,gsm04,gsm05,hyperrpca}. With the GSM prior, one can decompose $\boldsymbol{s}$ into point-wise product of a Gaussian vector $\boldsymbol{\alpha}$ and a positive hidden scalar multiplier $\boldsymbol{\theta}$ with probability $P\left(\theta_{j}\right)$, i.e., $s_{i}=\theta_{j}\alpha_{j}$. Conditioned on $\theta_{j}$, $s_{j}$ is Gaussian with standard deviation $\theta_{j}$, assuming that $\theta_{j}$ and $\alpha_{j}$ are independently identically distribution (IID), we can write the GSM prior of $\boldsymbol{s}$ as

  \begin{equation}
  \small
      P(\boldsymbol{s})=\prod_{j} P\left(s_{j}\right),
      P\left(s_{j}\right)=\int_{0}^{\infty} P\left(s_{j} | \theta_{j}\right) P\left(\theta_{j}\right) d \theta_{j}.
  \end{equation}

  Then, we compute the MAP estimation of $\boldsymbol{s}$ with the GSM prior by using a joint prior model $P\left(\boldsymbol{s},\boldsymbol{\theta}\right)$. By substituting $P\left(\boldsymbol{s},\boldsymbol{\theta}\right)$ into the MAP estimation of Eq. (\ref{equation.map}), we can obtain

  \begin{equation}\label{equation.estimation}
  \small
      (\boldsymbol{s}, \boldsymbol{\theta})=\mathop{\arg\min}_{\boldsymbol{s},\boldsymbol{\theta}}
      \{-\log P(\Tilde{\boldsymbol{s}} | \boldsymbol{s})-\log P(\boldsymbol{s} | \boldsymbol{\theta})-\log P(\boldsymbol{\theta})\}.
  \end{equation}

  Here we adopt the noninformative Jeffrey’s prior \cite{jeffreyprior}, i.e., $P\left(\theta_{j}\right)=1/\theta_{j}$, and Eq. (\ref{equation.estimation}) can be written into

  \begin{equation}\label{equation.Jeffrey}
  \small
      (\boldsymbol{s},\boldsymbol{\theta})=\mathop{\arg\min}_{\boldsymbol{s},\boldsymbol{\theta}}
      \|\Tilde{\boldsymbol{s}}-\boldsymbol{s}\|_{2}^{2} + \sigma_{w}^{2}\sum_{j}\frac{s_{j}^{2}}{\theta_{j}^{2}} +
      4\sigma_{w}^{2}\sum_{j}\log \theta_{j}.
  \end{equation}

  \noindent Note that in GSM we have $\boldsymbol{s}=\boldsymbol{\Lambda}\boldsymbol{\alpha}$, where $\boldsymbol{\Lambda}=\operatorname{diag}\left(\theta_{j}\right)\in\mathbb{R}^{N \times N}$. Then Eq. (\ref{equation.Jeffrey}) can be rewritten as

  \begin{equation}\label{equation.objective}
  \small
      \begin{split}
          (\boldsymbol{\alpha},\boldsymbol{\theta})= & \mathop{\arg\min}_{\boldsymbol{s},\boldsymbol{\theta}}
          \|\Tilde{\boldsymbol{s}}-\boldsymbol{\Lambda}\boldsymbol{\alpha}\|_{2}^{2} +
          \sigma_{w}^{2}\|\boldsymbol{\alpha}\|_{2}^{2} \\
          & + 4\sigma_{w}^{2}\sum_{j}\log\left(\theta_{j}+\epsilon\right),
      \end{split}
  \end{equation}

  \noindent where $\epsilon$ is a small positive constant. From Eq. (\ref{equation.objective}), we can see that the estimation of $\boldsymbol{s}$ has been translated into the joint estimation of $\boldsymbol{\alpha}$ and $\boldsymbol{\theta}$.

   \begin{algorithm}[t]
    \caption{Client Pruning Strategy}
    \label{algorithm.GSMpruning}
    \textbf{Require:} Active client set $\mathbb{U}_{t}$, $contribution\ scores$ $\Tilde{\boldsymbol{s}}^{t}$, $\sigma_{w}^{2}$ \\
    \textbf{Ensure:} Active client set $\mathbb{U}_{t+1}$
    \begin{algorithmic}[1]
        \STATE Initialize $\mathbb{U}_{t+1} \leftarrow \emptyset$.
        \WHILE{not converged}
        \STATE Compute $\boldsymbol{\theta}^{t}$ using Eq. (\ref{equation.theta}).
        \STATE Compute $\boldsymbol{\alpha}^{t}$ using Eq. (\ref{equation.alpha}).
        \STATE Compute $\boldsymbol{A}^{t}=\operatorname{diag}\left(\boldsymbol{\alpha}^{t}\right)$, $\boldsymbol{s}^{t}=\boldsymbol{A}^{t}\boldsymbol{\theta}^{t}$.
        \ENDWHILE
        \FOR{each client $k\in\mathbb{U}_{t}$}
        \STATE Set corresponding value of $k$ in $\boldsymbol{s}^{t}$ is $\boldsymbol{s}_{k}^{t}$.
        \IF{$\boldsymbol{s}_{k}^{t} > 0$}
        \STATE Add $k$ into $\mathbb{U}_{t+1}$ and remove it from $\mathbb{U}_{t}$.
        \ENDIF
        \ENDFOR
        \STATE \textbf{return} $\mathbb{U}_{t+1}$
    \end{algorithmic}
  \end{algorithm}

\subsection{Alternative Optimization}\label{subsection.Optimization}
  We propose an efficient algorithm solving Eq. (\ref{equation.objective}) by an alternating minimization, and we describe each sub-problem as follows.

  \subsubsection{Solving the \texorpdfstring{$\boldsymbol{\theta}$} {TEXT} Subproblem}
  When the $\boldsymbol{\alpha}$ is fixed, we can solve for $\boldsymbol{\theta}$ by optimizing

  \begin{equation}\label{equation.theta}
  \small
      \boldsymbol{\theta}=\mathop{\arg\min}_{\boldsymbol{\theta}}
      \|\Tilde{\boldsymbol{s}}-\boldsymbol{A} \boldsymbol{\theta}\|_{2}^{2} +
      4\sigma_{w}^{2}\sum_{j}\log\left(\theta_{j}+\epsilon\right),
  \end{equation}

  \noindent where $\boldsymbol{A}=\operatorname{diag}\left(\boldsymbol{\alpha}\right)$, and $\boldsymbol{\Lambda}\boldsymbol{\alpha}=\boldsymbol{A}\boldsymbol{\theta}$. Equivalenty, Eq. (\ref{equation.theta}) can be rewritten as

  \begin{equation}\label{equation.theta02}
  \small
      \boldsymbol{\theta}=\mathop{\arg\min}_{\boldsymbol{\theta}}
      \sum_{j}\left\{a_{j} \theta_{j}^{2}+b_{j} \theta_{j}+c \log \left(\theta_{j}+\epsilon\right)\right\},
  \end{equation}

  \noindent where $a_{j}=\alpha_{j}^{2}$, $b_{j}=-2\alpha_{j}\Tilde{s_{j}}$ and $c=4\sigma_{w}^{2}$. Thus, Eq. (\ref{equation.theta02}) boils down to solving a sequence of scalar minimization problems

  \begin{equation}\label{equation.theta03}
  \small
      \theta_{j}=\mathop{\arg\min}_{\boldsymbol{\theta_{j}}}
      a_{j}\theta_{j}^{2} + b_{j}\theta_{j} + c\log\left(\theta_{j}+\epsilon\right),
  \end{equation}

  \noindent which can be solved by taking $df(\theta_{j})/d\theta_{j}=0$, where $f(\theta)$ denotes the right hand side of Eq. (\ref{equation.theta03}). By taking $df(\theta_{j})/d\theta_{j}=0$, two stationary points can be obtained

  \begin{equation}\label{equation.theta04}
  \small
      \theta_{j,1}=-\frac{b_{j}}{4a_{j}}+\sqrt{\frac{b_{j}^{2}}{16}-\frac{c}{2a_{j}}},
      \theta_{j,2}=-\frac{b_{j}}{4a_{j}}-\sqrt{\frac{b_{j}^{2}}{16}-\frac{c}{2a_{j}}},
  \end{equation}

  \noindent then we denote $\Delta=b_{j}^{2} /\left(16 a_{j}^{2}\right)-c /\left(2 a_{j}\right)$, the solution to Eq. (\ref{equation.theta04}) can then be written as

  \begin{equation}\label{equation.theta05}
  \small
	\theta_{j}=
	\begin{cases}
	0&,\text{if $\Delta<0$} \\
	t_{j}&,\text{otherwise} \text{,}
	\end{cases}
  \end{equation}

  \noindent where $t_{j}=\mathop{\arg\min}_{\theta_{j}}\left\{f(0), f\left(\theta_{j, 1}\right), f\left(\theta_{j, 2}\right)\right\}$.

  \begin{algorithm}[t]
    \caption{FedCliP}
    \label{algorithm.fedcpl}
    \textbf{Server executes:}
    \begin{algorithmic}[1]
        \STATE Initialize global model $\boldsymbol{w}^{0}$, active client set $\mathbb{U}_{0} \leftarrow \mathbb{P}$.
        \FOR{each round $t=0,1, \ldots$}
        \STATE Select all clients $\boldsymbol{S}_{t}$ from $\mathbb{U}_{t}$.
        \FOR{each client $k\in\boldsymbol{S}_{t}$ \textbf{in parallel}}
        \STATE $\boldsymbol{w}_{k}^{t+1}, s_{k}^{t} \leftarrow \text {ClientUpdate}\left(k, \boldsymbol{w}^{t}\right)$.
        \ENDFOR
        \STATE $\boldsymbol{w}^{t+1} \leftarrow \sum_{k\in\boldsymbol{S}_{t}} p_{k}\boldsymbol{w}_{k}^{t+1}$.
        \STATE Collect select clients $contribution\ scores$ $\Tilde{\boldsymbol{s}}^{t}$.
        \STATE Update active client set $\mathbb{U}_{t+1} \leftarrow \mathbb{U}_{t}$ with Algorithm \ref{algorithm.GSMpruning}.
        \ENDFOR
    \end{algorithmic}
    \textbf{} \\
    \textbf{ClientUpdate}($k$, $\boldsymbol{w}$)\textbf{:}
    \begin{algorithmic}[1]
        \STATE Initialize the number of local epochs $E$, local minibatch size $B$, learning rate $\eta$.
        \STATE $\mathcal{B}\leftarrow$ (split $\mathbb{D}_{k}$ into batches of size $B$).
        \FOR{each local epoch $i$ from 1 to $E$}
        \FOR{batch $b\in\mathcal{B}$}
        \STATE $\boldsymbol{w}\leftarrow \boldsymbol{w}-\eta\nabla l_{k}\left(\boldsymbol{w};b\right)$
        \ENDFOR
        \ENDFOR
        \STATE Calculate $contribution\ score$ $s$ using Eq. (\ref{equation.clientrs}).
        \STATE \textbf{return} $\boldsymbol{w}$ and $s$ to server
    \end{algorithmic}
  \end{algorithm}

  \subsubsection{Solving the \texorpdfstring{$\boldsymbol{\alpha}$} {TEXT} Subproblem}
  For fixed $\boldsymbol{\theta}$, $\boldsymbol{\alpha}$ can be updated by solving

  \begin{equation}\label{equation.alpha}
  \small
      \boldsymbol{\alpha}=\mathop{\arg\min}_{\boldsymbol{\alpha}}
      \|\Tilde{\boldsymbol{s}}-\boldsymbol{\Lambda}\boldsymbol{\alpha}\|_{2}^{2} +
      \sigma_{w}^{2}\|\boldsymbol{\alpha}\|_{2}^{2},
  \end{equation}

  \noindent which also admits a closed-form solution namely

  \begin{equation}\label{equation.alpha02}
  \small
      \boldsymbol{\alpha}=\left(\boldsymbol{\Lambda}^{\top}\boldsymbol{\Lambda}+
      \sigma_{w}^{2}\boldsymbol{I}\right)^{-1}\boldsymbol{\Lambda}^{\top}\Tilde{\boldsymbol{s}},
  \end{equation}

  \noindent where $\boldsymbol{I}$ denotes the identity matrix. Since $\left(\boldsymbol{\Lambda}^{\top}\boldsymbol{\Lambda}+\sigma_{w}^{2}\boldsymbol{I}\right)$ is a diagonal matrix, Eq. (\ref{equation.alpha02}) can be easily computed.

  By alternatingly solving the sub-problems of Eq. (\ref{equation.theta}) and Eq. (\ref{equation.alpha}), sparse coefficients $\boldsymbol{s}$ can be estimated as $\hat{\boldsymbol{s}}=\hat{\boldsymbol{\Lambda}}\hat{\boldsymbol{\alpha}}$, where $\hat{\boldsymbol{\Lambda}}$ and $\hat{\boldsymbol{\alpha}}$ denote the estimates of $\boldsymbol{\Lambda}$ and $\boldsymbol{\alpha}$ respectively. At last, the clients corresponding to the non-zero values in $\boldsymbol{s}$ are the active clients when we obtain the active clients’ $contribution\ scores$ $\boldsymbol{s}$. In summary, the proposed GSM-based client pruning strategy is summarized in Algorithm \ref{algorithm.GSMpruning}, and the overall framework FedCliP based on Algorithm \ref{algorithm.GSMpruning} is summarized in Algorithm \ref{algorithm.fedcpl}. It is noteworthy that our method is orthogonal to existing FL optimizers that amend the training loss or the aggregation scheme, e.g., FedAvg \cite{fedavg} and FedProx \cite{fedprox}. So our method can be combined with any of them.

    \begin{figure*}[t]
	\centering
	\includegraphics[width=0.97\linewidth]{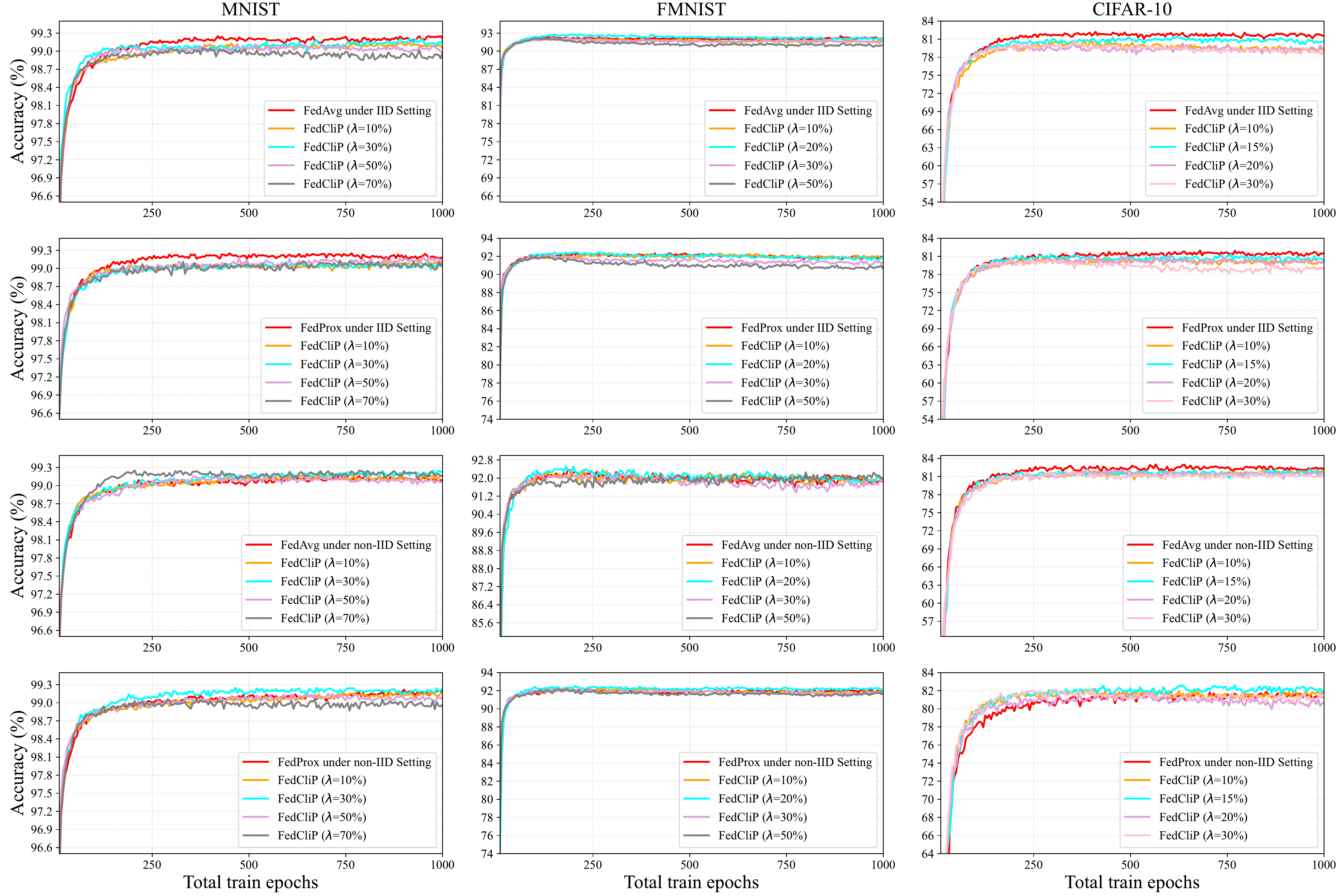}
	\caption{The test accuracy achieved by FedCliP on MNIST, FMNIST and CIFAR-10 under IID and non-IID settings.}
	\label{figure.accuracy}
  \end{figure*}

\section{Experiments}\label{section.Experiments}
\subsection{Experimental Setup}
  \textbf{Benchmark Datasets.} We conduct experiments on three classical datasets, including MNIST \cite{mnist}, FMNIST \cite{fmnist} and CIFAR-10 \cite{cifar10}. For MNIST, we adopt a CNN model with two convolutional layers followed by two linear layers. For FMNIST and CIFAR-10, we adopt the ResNet18 model with five convolutional layers followed by one linear layer. We partition each dataset into a training set and a test set with the same distributions, and transform the datasets according to IID distribution, and ensure the same distributions of local training and test set. We experiment with two different data partitions on N = 20 clients as follows.

  \noindent \textbf{IID and non-IID Settings.} We design two data distributions for empirical evaluation, IID setting and non-IID setting. For IID setting, each client distributes data equally to ensure consistent data distribution on each client. For non-IID setting, to simulate the data distribution more realistically, we randomly select 80\% of the data, distribute them to 10 clients according to the IID method. Then, we divide the remaining data into 20 shards ensuring that all the data in one shard have same label, and distribute them to the other 10 clients so that each client has two shards.

  \noindent \textbf{Baseline Studies.} We implement FedCliP following two state-of-the-art FL approaches to verify the feasibility and effectiveness of client pruning strategy: (1) FedAvg \cite{fedavg}, which takes the average of the locally trained copies of the global model; (2) FedProx \cite{fedprox}, which stabilizes FedAvg by including a proximal term in the loss function to limit the distance between the local model and last round global model.

  \noindent \textbf{Parameter Settings.} We train each method 200 communication rounds with 5 local epochs on each datasets. The client pruning ratio $\lambda$ on the MNIST datasets, we set $\lambda=$ 10\%, 30\%, 50\%, and 70\%, respectively. For FMNIST dataset, we set $\lambda=$ 10\%, 20\%, 30\% and 50\%, respectively. We set $\lambda=$ 10\%, 15\%, 20\% and 30\% on the CIFAR-10 dataset. During the experiment, due to the instability of model training in the first 20 communication rounds, we did not perform client pruning operations in the first 20 rounds and pruned a client in each training round after the first 20 rounds.

\subsection{Experimental Results under IID Setting}
  We summarize the empirical convergence behavior and performance under the IID and non-IID settings in Tables \ref{table.mnist}, \ref{table.fmnist}, \ref{table.cifar} and Figure \ref{figure.accuracy}.  The test accuracy curves of the total training rounds are shown in Figure \ref{figure.accuracy}, from which we can see that, under all experimental settings, all experimental training results can converge stably to a certain exact value, which means that using the FedCliP client pruning strategy can bring the federated training model to convergence.

  Corresponding to the IID setting, the FedCliP method performs best on the MNIST data set. Although the FedAvg and FedProx methods both achieve optimal performance without any client pruning, we can see from Table \ref{table.mnist} that when using FedCliP to dynamically prune clients during FL model training, even if the pruning rate is up to 70\%, the global model performance only drops by 0.2\%, still, the overall communication cost of the entire FL model has decreased by about 70\%. The experimental results on the FMNIST dataset show that as the client pruning rate increases, the overall performance of the global model shows a downward trend, which is a very intuitive and natural phenomenon. One can see from Table \ref{table.fmnist} that when the rate reaches 50\%, the model's performance only drops by 1\%, but the communication consumption at this time is reduced by 50\%.  As shown in Table \ref{table.cifar}, when we set the client pruning rate to be around 15\% to 20\%, we can control the model accuracy decrease of the FedCliP algorithm within 1\%, which means that the use of an acceptable model loss can maximize the reduction in communication consumption.

\subsection{Experimental Results under non-IID Setting}
  For non-IID setting, we find an interesting phenomenon in Table \ref{table.mnist}: the accuracy results of dynamically pruning 30\% of clients using the FedCliP strategy are the best on MNIST dataset, compared with the original method without any client pruning effects better. In the case of using all clients to participate in training, the data distributed on half of the clients is unbalanced, which accounts for 20\% of all train data and may harm the gradient descent of the model. Utilizing the method of dynamically pruning clients can effectively screen clients unfavorable to model convergence so that they do not continue to participate in the training of the global model, stabilizing the training results and minimizing communication overhead. The experimental results in Table \ref{table.fmnist} can also verify our statement. The performance of the FedProx using the client pruning strategy is higher than the original method without any client pruning. When the pruning rate reaches 50\%, the model performance is still slightly higher than the effect of not adopting the pruning strategy. The experimental results on the CIFAR-10 dataset in Table \ref{table.cifar} prove that the FedProx using the client pruning strategy can reduce the communication cost of FL without losing the model's accuracy.

\subsection{Ablation Study}
\begin{figure}[t]
	\centering
	\includegraphics[width=\linewidth]{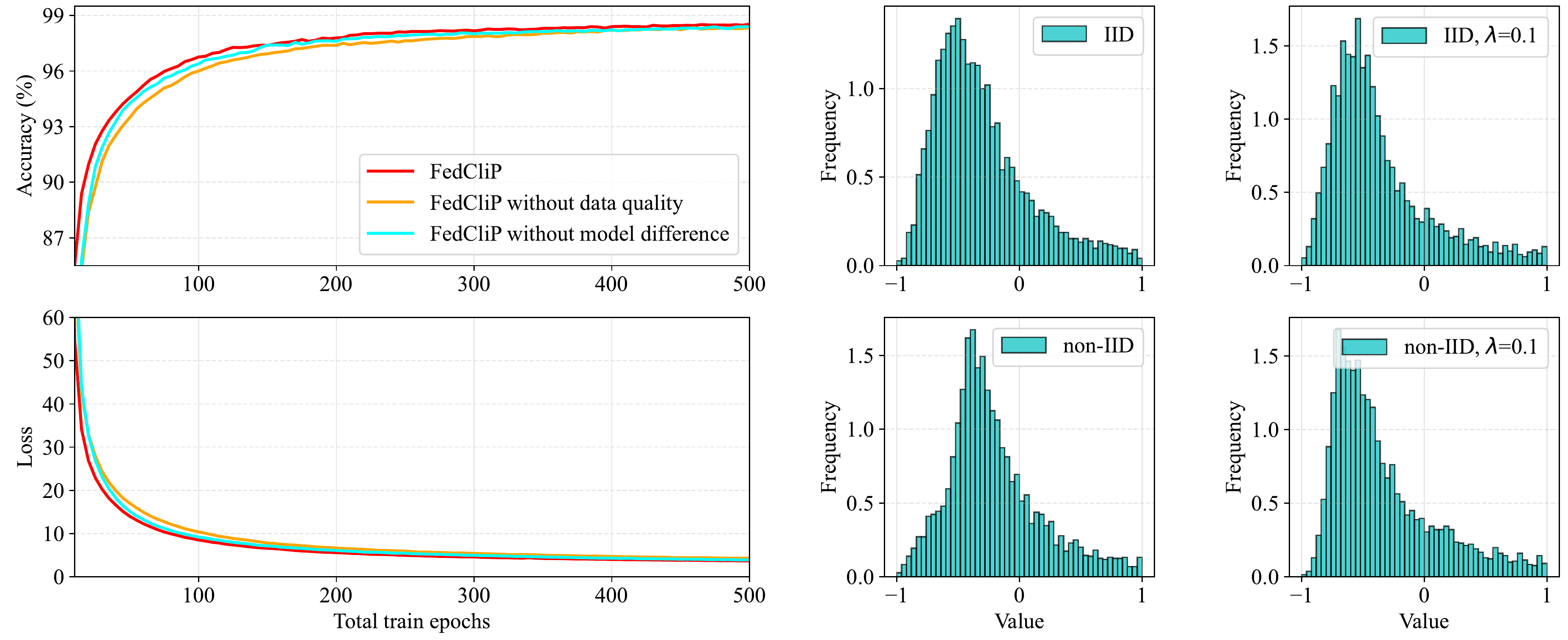}
	\caption{Left: the test accuracy and loss achieved by FedCliP ($\lambda=0.6$) on MNIST under IID setting. Right: the distributions of $contribution\ scores$ achieved by FedCliP on MNIST dataset.}
	\label{figure.ablation}
  \end{figure}

  The ablation experiments are designed to show the effectiveness of each component of $contribution\ score$ to the performance, ablation experiment results are shown in Figure \ref{figure.ablation}. The critical indicator of the FedCilP dynamic pruning client is the $contribution\ score$ calculated by each client after the current round of local training. The calculation of the $contribution\ score$ consists of two parts, the model difference, and the data quality. The experimental results in Figure \ref{figure.ablation} confirm the correctness of the pruning evaluation index we set. We only use model difference and data quality as a $contribution\ score$ standard, and the expected results can be obtained. We combine the two as the evaluation criteria for the FedCliP dynamic pruning client to improve the results.

\subsection{Results Discussion}
  As a novel solution to reduce the communication overhead, during the model training process, FedCliP gradually identifys active clients and prunes inactive clients. By setting up two more classic experimental scenarios, we verified the necessity and feasibility of gradually pruning the client during the FL process and reduced the communication overhead of the model as much as possible under the accuracy loss controllable. From IID setting, when we use FedCliP to dynamically prune clients during FL model training, even if the pruning rate is up to 70\%, the global model performance only drops by 0.2\%, still, the overall communication cost of the entire FL model has decreased by about 70\%. The non-IID setting can better reflect the performance superiority of the FedCliP algorithm. Due to the balanced distribution of some client data under the non-IID setting, it can easily lead to unstable model training. Through adaptive pruning clients during the model training process, accurately pruning inactive clients and retaining active clients can improve the stability of model training. Utilizing the method of dynamically pruning clients can effectively screen clients unfavorable to model convergence so that they do not continue to participate in the training of the global model, stabilizing the training results and minimizing communication overhead.

  \begin{table}[t]
  \centering
  \renewcommand\arraystretch{1.1}
  \resizebox{\linewidth}{!}{
    \begin{tabular}{clccc}
    \hline
    \multicolumn{2}{c}{\multirow{2}{*}{Method}}                                     & \multicolumn{2}{c}{Accuracy} & \multirow{2}{*}{\begin{tabular}[c]{@{}c@{}}Communication costs\end{tabular}} \\ \cline{3-4}
    \multicolumn{2}{c}{}                                                            & IID         & non-IID        &                                                                                \\ \hline
    \multicolumn{2}{l}{FedAvg}                                                      &$\boldsymbol{99.25}$        &99.16           &1.00                                                             \\ \hdashline
    \multirow{4}{*}{\begin{tabular}[c]{@{}c@{}}FedCliP\end{tabular}}         & 10\% &99.14        &99.18           &0.90                                                                            \\
                                                                             & 30\% &99.21        &$\boldsymbol{99.25}$           &0.70                                                             \\
                                                                             & 50\% &99.11        &99.15           &0.50                                                                            \\
                                                                             & 70\% &99.05        &99.25           &0.30                                                                            \\ \hline
    \multicolumn{2}{l}{FedProx}                                                     &$\boldsymbol{99.25}$        &99.22           &1.00                                                             \\ \hdashline
    \multirow{4}{*}{\begin{tabular}[c]{@{}c@{}}FedCliP\end{tabular}}         & 10\% &99.14        &99.19           &0.90                                                                            \\
                                                                             & 30\% &99.12        &$\boldsymbol{99.25}$           &0.70                                                             \\
                                                                             & 50\% &99.19        &99.15           &0.50                                                                            \\
                                                                             & 70\% &99.13        &99.06           &0.30                                                                            \\ \hline
    \end{tabular}
  }
  \caption{Best test accuracy (\%) achieved by FedCliP on MNIST dataset under IID and non-IID settings. Highest performance is represented in bold.}
  \label{table.mnist}
  \end{table}

  \begin{table}[t]
  \centering
  \renewcommand\arraystretch{1.1}
  \resizebox{\linewidth}{!}{
    \begin{tabular}{clccc}
    \hline
    \multicolumn{2}{c}{\multirow{2}{*}{Method}}                                     & \multicolumn{2}{c}{Accuracy} & \multirow{2}{*}{\begin{tabular}[c]{@{}c@{}}Communication costs\end{tabular}} \\ \cline{3-4}
    \multicolumn{2}{c}{}                                                            & IID         & non-IID        &                                                                                \\ \hline
    \multicolumn{2}{l}{FedAvg}                                                      &92.50        &92.36           &1.00                                              \\ \hdashline
    \multirow{4}{*}{\begin{tabular}[c]{@{}c@{}}FedCliP\end{tabular}}         & 10\% &92.05        &92.32           &0.90                                                                            \\
                                                                             & 20\% &$\boldsymbol{92.73}$        &$\boldsymbol{92.37}$           &0.80                                                                            \\
                                                                             & 30\% &92.05        &92.20           &0.70                                                                            \\
                                                                             & 50\% &91.51        &92.25           &0.50                                                                            \\ \hline
    \multicolumn{2}{l}{FedProx}                                      &$\boldsymbol{92.43}$        &92.24           &1.00                                                             \\ \hdashline
    \multirow{4}{*}{\begin{tabular}[c]{@{}c@{}}FedCliP\end{tabular}}         & 10\% &92.37        &92.27           &0.90                                                             \\
                                                                             & 20\% &92.43        &$\boldsymbol{92.43}$           &0.80                                                                            \\
                                                                             & 30\% &91.76        &92.16           &0.70                                                                            \\
                                                                             & 50\% &91.54        &92.02           &0.50                                                                            \\ \hline
    \end{tabular}
  }
  \caption{Best test accuracy (\%) achieved by FedCliP on FMNIST dataset under IID and non-IID settings. Highest performance is represented in bold.}
  \label{table.fmnist}
  \end{table}

  \begin{table}[t]
  \centering
  \renewcommand\arraystretch{1.1}
  \resizebox{\linewidth}{!}{
    \begin{tabular}{clccc}
    \hline
    \multicolumn{2}{c}{\multirow{2}{*}{Method}}                                     & \multicolumn{2}{c}{Accuracy} & \multirow{2}{*}{\begin{tabular}[c]{@{}c@{}}Communication costs\end{tabular}} \\ \cline{3-4}
    \multicolumn{2}{c}{}                                                            & IID         & non-IID        &                                                                                \\ \hline
    \multicolumn{2}{l}{FedAvg}                                                      &$\boldsymbol{82.22}$        &$\boldsymbol{83.01}$           &1.00                                              \\ \hdashline
    \multirow{4}{*}{\begin{tabular}[c]{@{}c@{}}FedCliP\end{tabular}} & 10\% &80.92        &82.04           &0.90                                                                            \\
                                                                             & 15\% &81.37        &82.21           &0.85                                                                            \\
                                                                             & 20\% &80.32        &82.14           &0.80                                                                            \\
                                                                             & 30\% &80.66        &81.87           &0.70                                                                            \\ \hline
    \multicolumn{2}{l}{FedProx}                                                     &$\boldsymbol{81.98}$        &81.88           &1.00                                                             \\ \hdashline
    \multirow{4}{*}{\begin{tabular}[c]{@{}c@{}}FedCliP\end{tabular}} & 10\% &80.78        &82.24           &0.90                                                                            \\
                                                                             & 15\% &81.39        &$\boldsymbol{82.60}$           &0.85                                                             \\
                                                                             & 20\% &81.05        &81.83           &0.80                                                                            \\
                                                                             & 30\% &80.43        &82.22           &0.70                                                                            \\ \hline
    \end{tabular}
  }
  \caption{Best test accuracy (\%) achieved by FedCliP on CIFAR-10 dataset under IID and non-IID settings. Highest performance is represented in bold.}
  \label{table.cifar}
  \end{table}

\section{Conclusion and Future Work}\label{section.Conclusion}
  In this paper, we propose a novel FL framework for communication efficiency, named FedCliP, by adaptively learning which clients should remain active for next round model training and pruning those clients that should be inactive and have less potential contribution. We also introduced an alternative optimization method based on GSM modeling, using a newly defined $contribution\ score$ metric to help identify active and inactive clients. We empirically evaluate the communication efficiency of the FL framework by conducting extensive experiments on three benchmark datasets under IID and non-IID settings. Numerical results show that the proposed FedCliP framework outperforms the state-of-the-art FL framework for efficient communication, i.e., FedCliP can save 70\% of communication overhead with only 0.2\% accuracy loss on MNIST dataset, and save 50\% on FMNIST and 15\% on CIFAR-10 communication overhead, respectively, with an accuracy loss of less than 1\%. In our future work, we will consider the client recycling mechanism to enhance the performance of our proposed algorithm further.

\bibliographystyle{named}
\bibliography{ijcai23}







\end{document}